\documentclass{article}

\usepackage[final, nonatbib]{neurips_2020}

\usepackage[utf8]{inputenc} 
\usepackage[T1]{fontenc}    
\usepackage{hyperref}       
\usepackage{url}            
\usepackage{booktabs}       
\usepackage{amsfonts}       
\usepackage{nicefrac}       
\usepackage{microtype}      
\usepackage{mathtools}
\usepackage{graphicx}
\usepackage{algorithm}
\usepackage[noend]{algpseudocode}

\title{Learning Principle of Least Action with Reinforcement Learning}

\author{%
  Zehao Jin\thanks{equal contribution} \\
  New York University Abu Dhabi \\
  \texttt{zj448@nyu.edu } \\
  \AND
  Joshua Yao-Yu Lin$^*$ \\
  Department of Physics\\
  University of Illinois at Urbana-Champaign\\
  \texttt{yaoyuyl2@illinois.edu} \\
  \And
  Siao-Fong Li \\
  University of Massachusetts, Amherst \\
  \texttt{siaofongli@umass.edu} \\

}

\begin{document}

\maketitle

\begin{abstract}



Nature provides a way to understand physics with reinforcement learning since nature favors the economical way for an object to propagate. In the case of classical mechanics, nature favors the object to move along the path according to the integral of the Lagrangian, called the action $\mathcal{S}$. We consider setting the reward/penalty as a function of $\mathcal{S}$, so the agent could learn the physical trajectory of particles in various kinds of environments with reinforcement learning. In this work, we verified the idea by using a Q-Learning based algorithm on learning how light propagates in materials with different refraction indices, and show that the agent could recover the minimal-time path equivalent to the solution obtained by Snell's law or Fermat's Principle. We also discuss the similarity of our reinforcement learning approach to the path integral formalism.

\end{abstract}

\section{Introduction}

\begin{figure}[htbp]
  \centering
  \includegraphics[width=1\linewidth]{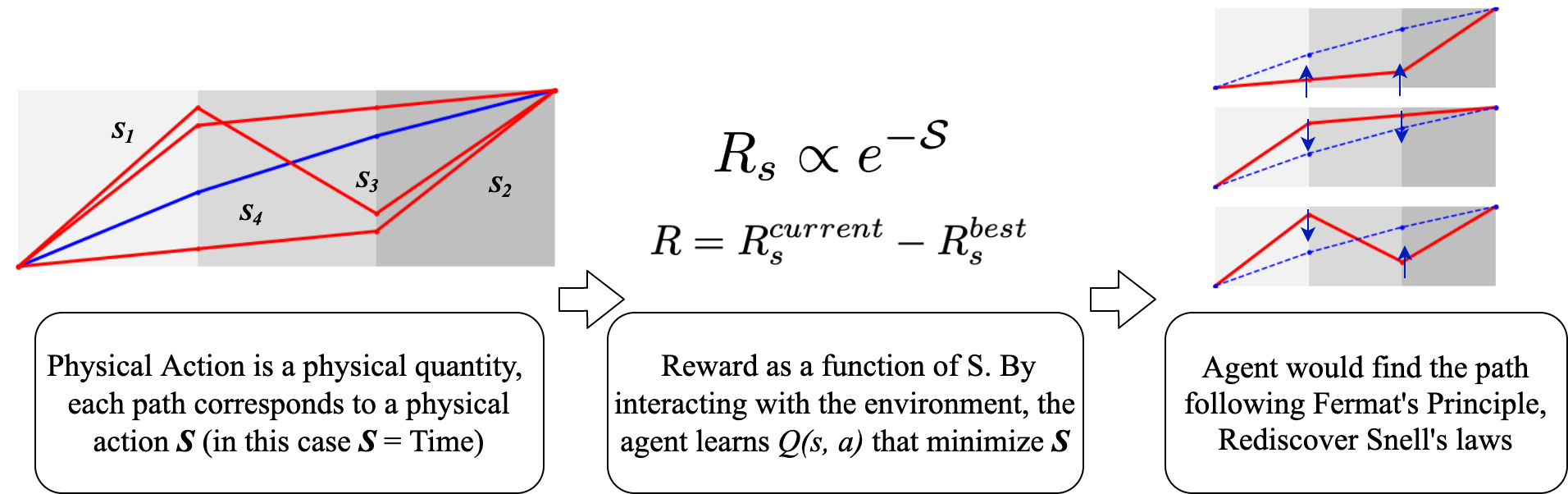}
  \caption{A cartoon summarizing how we connect the principle of least action and reinforcement learning in the case of light refraction.}
  \label{fig:RL_demo}
\end{figure}

Understanding the nature of how an object moves has been a fundamental question in physics.  Newtonian mechanics (e.g. Newton’s three laws of motion) have been very successful in describing the equation of motion in a classical world. Newtonian mechanics requires considering all the constraint forces to obtain the equation of motion. On a different approach, the Lagrangian mechanics provides a unique perspective that avoids the considerations of the constraint forces, and the Euler-Lagrangian has proved to be an easier way to solve the equation of motion of complex systems such as multiple pendula. The central idea of Euler-Lagrangian is the principle of least action. The principle of least action is the basic variational principle of particle and continuum systems: a trajectory of a system between initial and final states in a specified time is found by imagining all possible trajectories within that system, computing the action $\mathcal{S}$ for each of these trajectories, and selecting one that makes the action locally stationary (traditionally called "least"). True trajectories are those that have the least action.


Several studies have been done using deep neural networks with inductive bias that incorporate the Euler-Lagrange equation to predict the motion of objects. Most of them are done in a supervised way \cite{lutter2019deep, cranmer2020lagrangian, greydanus2019hamiltonian, toth2019hamiltonian}. These could be restrictive because the training set of the system of interest needs to be designed exactly as the testing set, while in reality, the governing dynamics of the task of interest are usually unknown or only partial information is revealed. Here we proposed that we could use reinforcement learning to learn the dynamics of a physical system.  


Reinforcement Learning (RL) has been recently showing promising progress in various applications, especially when applied to games \cite{mnih2013playing, silver2016mastering}, robotic  \cite{kalashnikov2018qt} and scientific discovery or design \cite{halverson2019branes, garnier2019review, popova2018deep, denil2016learning}. In RL, an agent in an unknown environment would need to explore and collect rewards by interacting with the environment. In this work, we aim to use the RL algorithm to solve problems in classical physics. 


We used the refraction of light combined with Q-Learning as an example to demonstrate the concept \cite{watkins1989learning}. Light rays (agent) travel in different materials (environment) and we calculate the time it took (reward function) for the path. We set up our starting point $A$ and final point $B$. We also restrict the light rays to move in a straight line in a material. We set that the agent must end up at $B$ point, so that our agent would always reach the destination in each round within an episode. The light path used to calculate the reward in each round is just straight line segments connecting starting point $A$, incident points at different interfaces, and final point $B$. Our agent will then search along with the interface (RL action) for the location of incident points (RL states) at each interface that gives the light path of the shortest time. That is, the agent will always get some reward in each round depending on its choice of incident points in each material.



We noticed the word \textit{action} is used in a different way in physics and RL. In this work we use "physical action" $\mathcal{S}$ to represent the physical quantity, not to be confused with the RL action $a$. 


\section{Reward as a function of Physical Action $\mathcal{S}$}

\subsection{Fermat's principle of least time}

In order to determine the time the photon spends between two points A and B, we could integrate the time $dt$ it spend at every instance, which would be distance divided by its ray's velocity in the media, $\mathcal{T} = \int_{A}^B dt = \int_{A}^B \frac{ds}{v_r}$, where $v_r$ represents the speed of light in the medium. Fermat's principle states that the light path between two given points would be the one traversed in the least time. We set the speed of light $c = 1$, then this could be stated as: $\delta \mathcal{S} ~ \text{(optical path length)} = \delta \mathcal{T} = \delta \int_A^{B} n_r ds =  0$, where $n_r$ represents the refraction index. We further generalized it to optical path length that would have the form of (physical) action \cite{chaves_2017}. Once we have the physical action, we could choose a reward as a function of the physical action, in this case, a function of time as the reward.








\begin{figure}[htbp]
  \centering
  \includegraphics[width=1\linewidth]{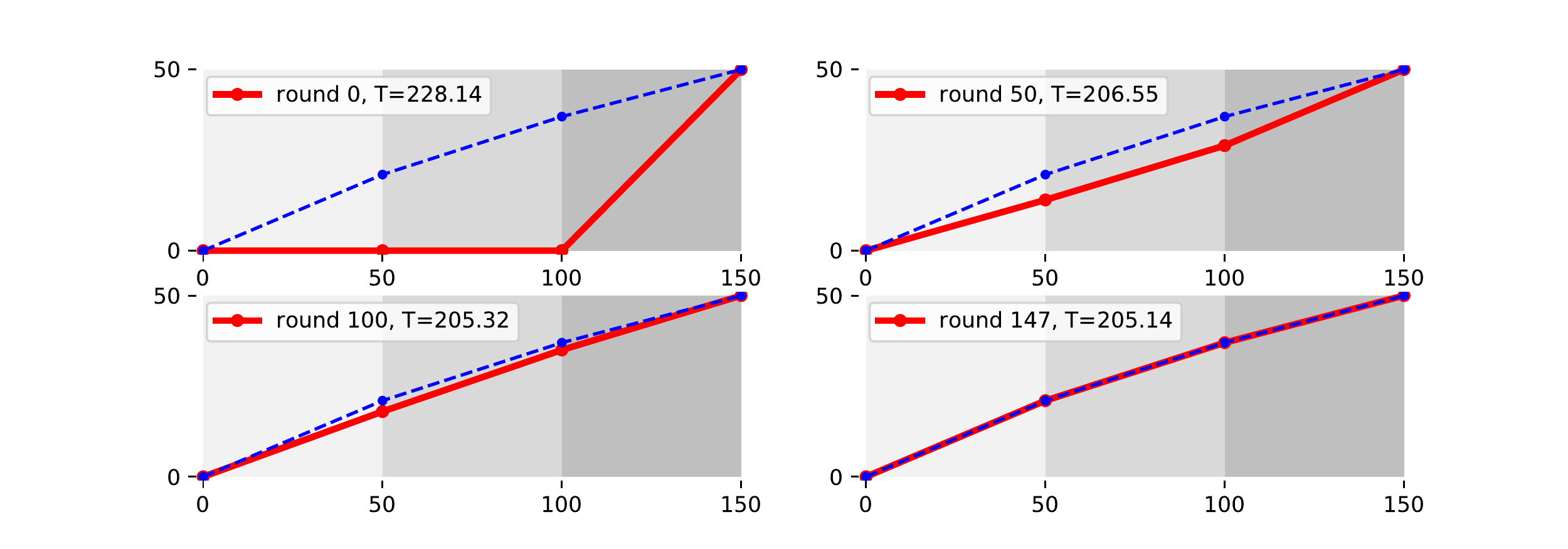}
  \caption{The evolution of learning during a single training episode. The environment of air($n_{air}=1$), water($n_{water}=1.3$), and glass($n_{glass}=1.6$) from left to right. Our agent is asked to travel from bottom left corner to top right corner. The red lines are the light path our agent chose and the blue dotted line is the theoretical least-time light path.}
  \label{fig:steps}
\end{figure}

\subsection{Experiment}

We experiment with our idea upon a simple case, the refraction of light, with the Q-Learning. An overview of our experiment is described in the table and visualized in Figure \ref{fig:steps}.

Q-Learning \cite{WatkinsDayan} is a basic RL algorithm that builds a Q-table that keeps record of Q-values of all available actions at all possible states. The Q-value is updated based on the reward the agent received by making a particular action at a particular state. At any state, there is an $\epsilon$ chance for the agent to make an action that has the highest Q-value, otherwise the agent will take a random action for more exploration. In this work we adopted the greedy factor $\epsilon=0.9$, learning rate $\alpha$ = 0.001, discount factor $\gamma$ = 0.9.

Based on Fermat's principle, the time $T$ the photon takes to travel through some distance $l$ in a material with index of refraction $n_r$ is $T=l \cdot n_r$. As light travels through multiple different materials, the total time it takes is just the sum of time took in each material, $T = \Sigma_i T_i$, and $T_i = l_{i} \cdot n_{r, i}$.

\begin{tabular}{ |p{4cm}|p{1cm}| p{7.5cm}| }
 \hline
 \multicolumn{3}{|l|}{Q-Learning and Environment General parameters} \\
 \hline
 RL State & $s$ & $(y_1, y_2)$, with $0 \leq y_1 \leq 50$   and $0 \leq y_2 \leq 50$\\
 \hline
 RL Action & $a$ & $y1 \uparrow, y1 \downarrow, y2 \uparrow, y2 \downarrow$  \\
 \hline
 RL Reward   & $R$ &  $R=R_s^{\textrm{current}} - R_s^{\textrm{best}}$, $R_s= N e^{-T}$\\ 
 \hline
 Total Time & $T$ & $T = \Sigma_i T_i, ~~~T_i = l_{i} \cdot n_{r, i}$ \\
 \hline
 Greedy factor & $\epsilon$ &  $\epsilon=0.9$\\ 
 \hline
 Learning rate & $\alpha$ &  $\alpha = 0.001$\\ 
 \hline
 Discount factor & $\gamma$ &  $\gamma = 0.9$\\ 
 \hline
 \multicolumn{3}{|l|}{Parameters specifically for the case in Figure \ref{fig:steps}} \\
 \hline
 Index of refraction & $n_i$ & left to right: $n_{air}=1$, $n_{water}=1.3$, $n_{glass}=1.6$\\
 \hline
 Path endpoints & $A,B$ & $A(x,y)=(0,0), B(x,y)=(150,50)$ \\
 \hline
 Initial state & $s_{ini}$ & $s_{ini}(y_1,y_2)=(0,0)$ \\
 \hline
 Snell's Law prediction & $s_{theo}$ & $s_{theo}(y_1,y_2)=(21,37)$  \\
 \hline
 Total training episode   &   &  100\\ 
 \hline
 Rounds in each episode   &   &  300\\ 
 \hline

\end{tabular}

We constructed a three-layer, $50 \times 150$ grid environment that consists of three $50 \times 50$ grid materials of air, water and glass from left to right. The given endpoints for our RL agent is from bottom left corner($A$) to top right corner($B$).

The RL state of our agent is State=($y_1$-coordinate of air-water incident point,$y_2$-coordinate of water-glass interface). At the beginning of each training episode, our agent starts from initial state $s_{ini}(y_1,y_2)=(0,0)$, and the theoretical least-time light path is state $s_{theo}(y_1.y_2)=(21,37)$. Each round our agent moves up/down one unit along one of the two interfaces. That is, for each round, the one of the four RL actions $a=\{y1 \uparrow, y1 \downarrow, y2 \uparrow, y2 \downarrow\}$ is taken, where the arrow means moving along the direction for one unit.

We defined a R score, $R_s$ as
\begin{equation}
    R_s= N e^{-\mathcal{S}}  = N e^{-T} 
\end{equation}


where $T$ is the total time it takes for our agent to travel between two endpoints, and $N$ is just an arbitrary scaling factor. The $e^{-\mathcal{S}}$ form is taken from the Euclidean path integral formalism \cite{hall2013quantum}. The reward our agent receives for each round is the difference between $R_s$ for this round and the best $R_s$ achieved so far in this episode.
\begin{equation}
    R=R_s^{\textrm{current}} - R_s^{\textrm{best}}
\end{equation}
The reward is defined this way so that the agent is would get reward if the current path is better than the path it has explored, and vice versa. We find that in this particular environment setting, this definition of reward and $R_s$ help the agent find a global maximum faster.

\begin{figure}
  \centering
  \includegraphics[width=0.49\linewidth]{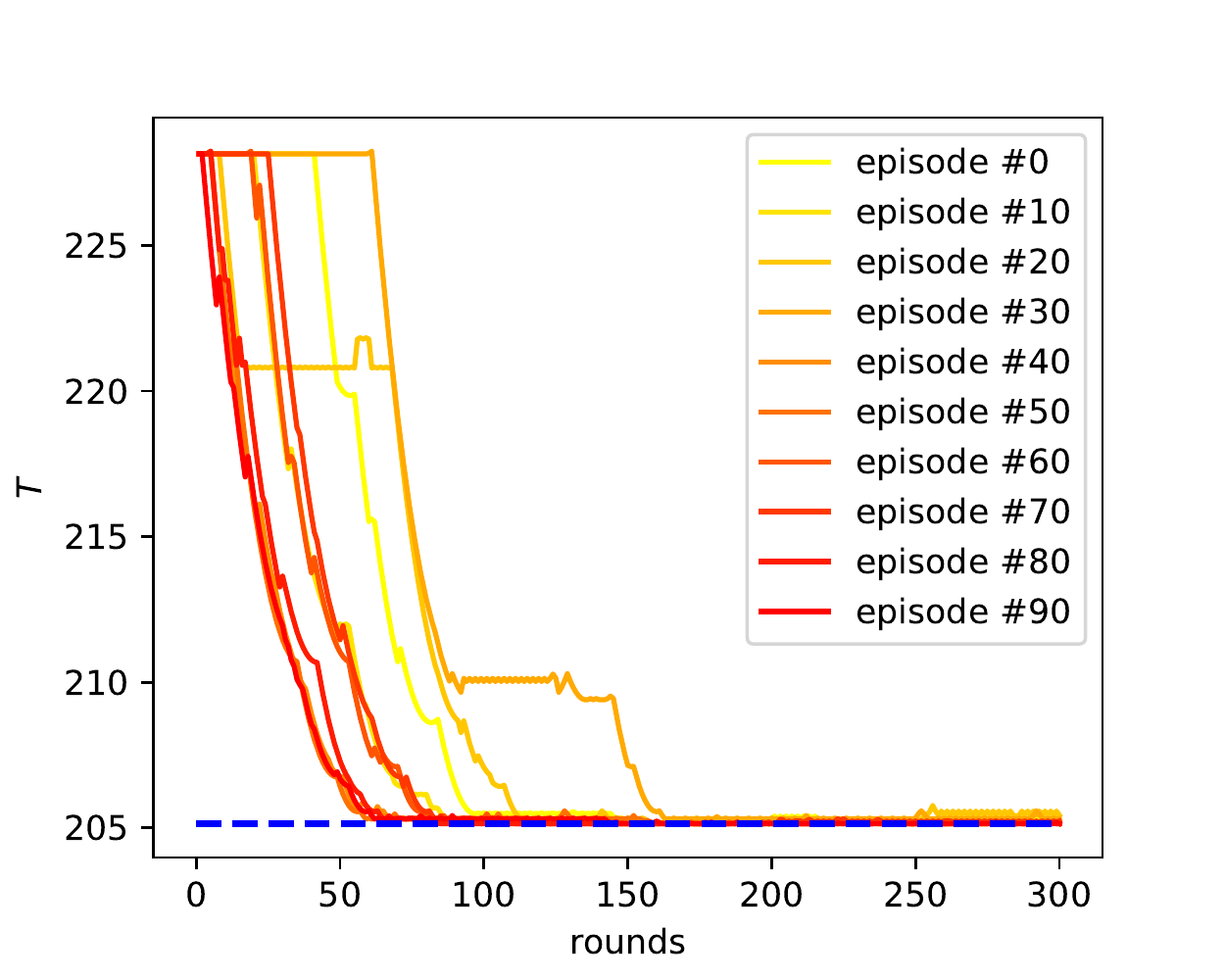}
  \includegraphics[width=0.49\linewidth]{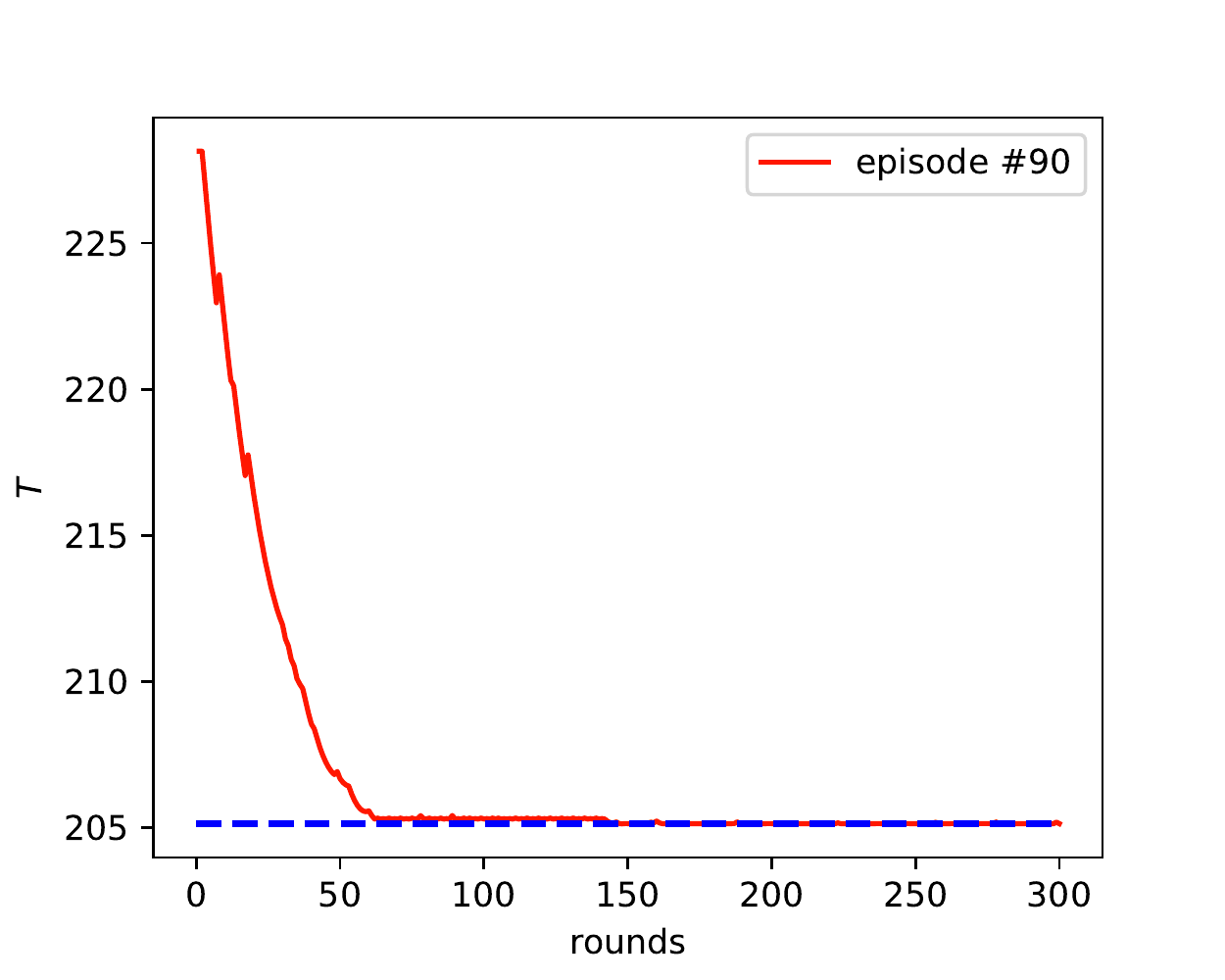}
  \caption{Left panel shows the time our agent took to travel between the given endpoint for each training round during 10 different training episodes. The red dotted line denotes the theoretical least-time light path. Right panel gives a clearer view for episode $\#$90, which is taken as an example in Figure \ref{fig:steps}.}
  \label{fig:episodes}
\end{figure}

Our agent is trained for 100 episodes, and during each episode our agent moves 300 rounds. The training result for each episode can be visualized in Figure \ref{fig:episodes}, and as an example, Figure \ref{fig:steps} visualizes training episode $\#$90. Our agent is able to find the path that takes the least time.

\begin{figure}[htbp]
  \centering
  \includegraphics[width=1\linewidth]{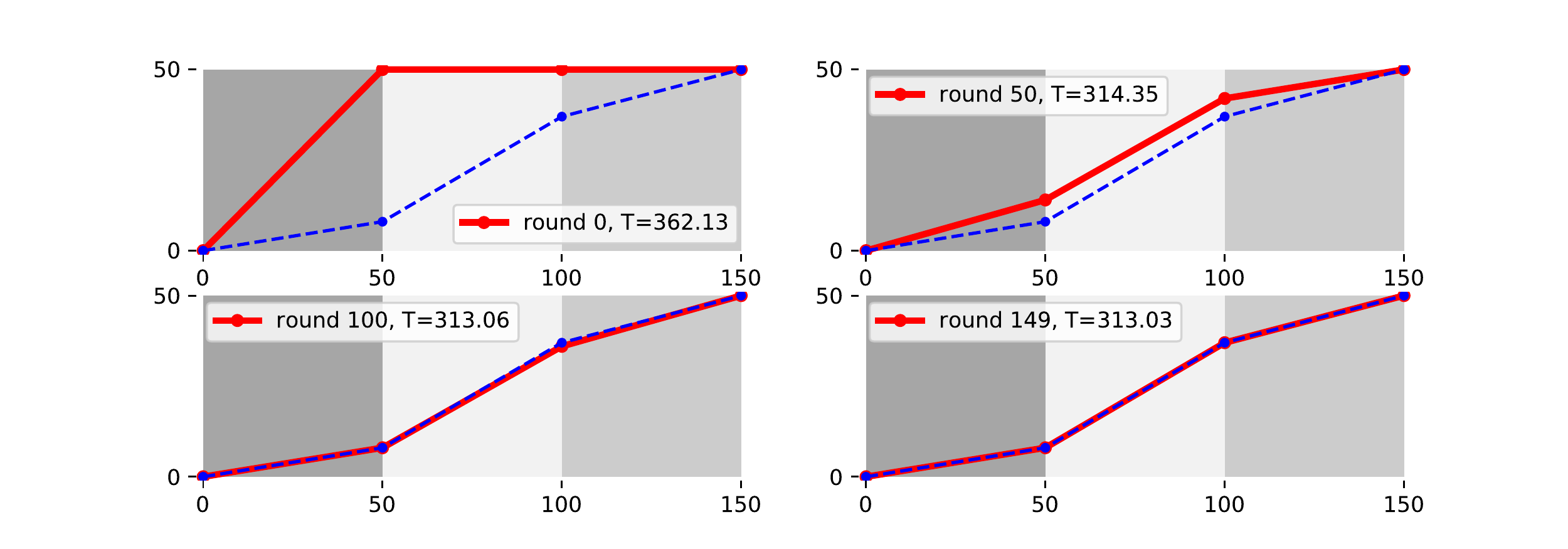}
  \caption{The environment of refraction index $n_1,n_2,n_3=(3,1,2)$ from left to right. The search begins from initial state $s_{ini}(y_1,y_2)=(50,50)$. The red lines are the light path our agent chose during episode $\#$90, and the blue dotted line is the theoretical least-time light path.}
  \label{fig:steps2}
\end{figure}

To generalize the problem a bit,the agent is also trained under indices of refraction other than $n_1,n_2,n_3=(1,1.3,1.6)$, and with initial states other than $s_{ini}(y_1,y_2)=(0,0)$. The agent can still successfully find the correct path. One of those trials are shown in Figure \ref{fig:steps2}.

\section{Discussion}

We demonstrate that RL can be applied to path finding in a physical system by purely interacting with the environment and getting the path of least action without human guidance or prior physics knowledge. We believe that with more computational resources and advanced RL algorithm, RL can be applied to solve complex environments to help us solve physics problems. We also noticed that the \textit{exploration} and \textit{exploitation} nature of RL approach is actually similar to the spirit of path integrals in quantum mechanics and quantum field theory \cite{dirac1981principles}, where all of the conceivable (non-optimized) path could also contribute so the all possible path between two points needs to be explored and evaluated. We plan to investigate such ideas in our future work.








\section*{Broader Impact}

We plan to open-source our codes along with publication. For the physics community, we demonstrate that a general algorithm could learn the trajectory/ equation of motion of the physical system. Also, we believe that the RL approach is similar to the path integral formalism so our work could stimulate research for quantum physics and reinforcement learning to help us better understand fundamental physics. We hope that this work serves as an effort to combine research in the reinforcement learning and physics community.


\begin{ack}
The authors thanks Miles Cranmer, Ji Won Park, Gabriella Contardo, Shirley Ho, Ashley Villar, Rodin Luo, Harrey Lee, Cathy Shih, Adam Liu, Jian Peng, and Zhizhen Zhao for useful discussion. JL thanks the AWS Cloud Credits for Research program.

\end{ack}

\small
\bibliographystyle{unsrt}
\bibliography{references}

\end{document}